\documentclass{article}
\usepackage[normalem]{ulem}
\useunder{\uline}{\ul}{}
\usepackage{enumitem}
\usepackage{array}
\usepackage{cite}
\usepackage{tabularray}
\usepackage{stfloats}
\usepackage{diagbox}
\usepackage{multirow}
\usepackage{amssymb}
\usepackage{graphicx}
\usepackage[hidelinks]{hyperref}
\usepackage[numbers]{natbib}
\usepackage{spconf,amsmath,epsfig}

\let\OLDthebibliography\thebibliography
\renewcommand\thebibliography[1]{
  \OLDthebibliography{#1}
  \setlength{\parskip}{0pt}
  \setlength{\itemsep}{0pt plus 0.3ex}
}

\begin{document}\sloppy

% Example definitions.
% --------------------

% Title.
% ------
\title{Trinity Detector:text-assisted and attention mechanisms based spectral fusion for diffusion generation image detection}
%
% Single address.
% ---------------
\name{Jiawei Song, Dengpan Ye, Yunming Zhang}
%Address and e-mail should NOT be added in the submission paper. They should be present only in the camera ready paper. 
\address{}

\maketitle

\begin{abstract}
Artificial Intelligence Generated Content (AIGC) techniques, represented by text-to-image generation, have led to a malicious use of deep forgeries, raising concerns about the trustworthiness of multimedia content. Adapting traditional forgery detection methods to diffusion models proves challenging. Thus, this paper proposes a forgery detection method explicitly designed for diffusion models called \textbf{Trinity Detector}. Trinity Detector incorporates coarse-grained text features through a CLIP encoder, coherently integrating them with fine-grained artifacts in the pixel domain for comprehensive multimodal detection. To heighten sensitivity to diffusion-generated image features, a Multi-spectral Channel Attention Fusion Unit \textbf{(MCAF)} is designed, extracting spectral inconsistencies through adaptive fusion of diverse frequency bands and further integrating spatial co-occurrence of the two modalities. Extensive experimentation validates that our Trinity Detector method outperforms several state-of-the-art methods,  our performance is competitive across all datasets and up to 17.6\% improvement in transferability in the diffusion datasets.
\end{abstract}
\section{Introduction}
\label{sec:intro}
Recently, diffusion models have rapidly advanced the field of image generation. AI generation technologies, exemplified by text-image generation, have significantly reduced the barriers to synthetic image creation. Unfortunately, this capability has the potential to be abused for malicious purposes. For instance, text-image generation can be utilized in zero-shot scenarios to craft deepfake attacks targeting prominent political figures worldwide~\cite{zhu_data_2023}. This misuse can potentially engender severe trust issues within our societal fabric. The diffusion generation mechanism differs from previous approaches, and existing detection methods exhibit poorly in its transferability. Thus, it is of high significance to develop a forgery detection method for diffusion models.

There have been some recent research on image detection for diffusion model generation, some research~\cite{wang_cnn-generated_2020}~\cite{tan2023rethinking} employs convolutional neural networks trained on different datasets for extracting features and performing classification. Zhong et al.~\cite{zhong_rich_2023} proposed a detection model that extracts the difference features by analyzing the pixel correlation between rich texture and weak texture regions in the image, but only considering the single feature of texture leads to poor data generalization, etc. Huang et al.~\cite{10219991} proposed to incorporate image-text bimodality into the detection of false content by stacking cross-aligned bimodal features, but this did not achieve alignment in semantic space.

\begin{figure}
  \centering
  \includegraphics[width=0.5\textwidth]{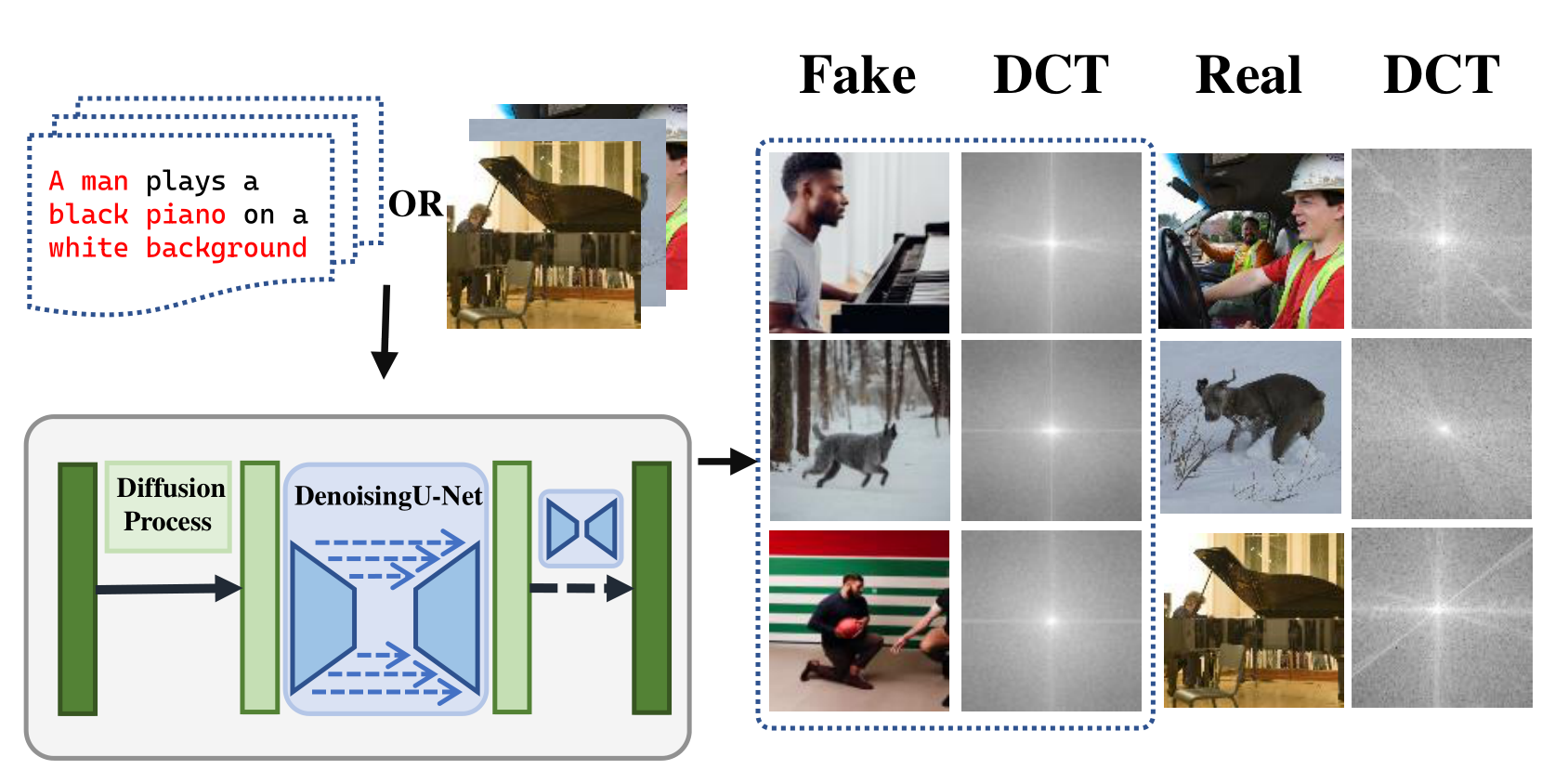} 
  \caption{\textbf{Diffusion generation process and comparison of spectrogram of real and fake images after DCT transformation. } }
  \label{fig:fre}
\end{figure}

To address this challenge, our work analyzes the differential representation of diffusion models and real images. It has been found that the prompt input has a significant impact on whether the generated image is more realistic or not~\cite{sha2023fake}, the semantic coarseness of textual prompts and the structural integrity of its semantic space have a direct impact on the quality of diffusion model generation, and it can be expected that the relevant characteristics of the text also have a feature-level mapping impact on the detection of images generated by the diffusion model. Given that U-Net has been widely adopted in various diffusion models, the up-sampling layer included in U-Net has also become a vital component of the diffusion model, and the frequency-domain content changes brought about by the up-sampling layer also have a great potential to be used as a generalized diffusion forgery detection. As shown in Fig~\ref{fig:fre},  through the frequency domain transformation, we find that the frequency domain of the real image is more balanced in all directions, while the diffusion-generated image is concentrated in specific directions.

We propose a new method called Trinity Detector. Trinity Detector provides a reliable way to distinguish between real images and diffusion-generated images. Precisely, the Trinity Detector method consists of two parts: (1) We specially design a multispectral channel attention fusion unit to extract the spectral inconsistencies between real images and diffusion model-generated images through adaptive fusion of different frequency bands to further incorporate the spatial co-occurrence of these two modalities. (2) Semantic spatial alignment fusion by introducing coarse-grained features of textual information with fine-grained artifacts in the pixel domain through the CLIP encoder.

Additionally, to train and evaluate the diffusion-generated image detector, we created a comprehensive diffusion-generated dataset, which includes images generated by training on Stable Diffusion and GLIDE.

In summary, our major contributions of this work are three-fold as follows:

\begin{itemize}[itemsep=1pt,topsep=2pt,parsep=0pt]
% \begin{itemize}[]
\item We propose a new forgery detection method called Trinity Detectorthat, it uses an attention mechanism to fuse the frequency domain and text-assisted visual content for diffusion image forgery detection.
\item We propose a Multi-spectral Channel Attention Fusion Unit (MCAF), that extracts the spectral inconsistencies between real images and images generated by the diffusion model through adaptive fusion across different frequency bands.
\item We produced a diffusion model-based image-text pair dataset for benchmarking the diffusion-generated image detectors.
\item Extensive experiments show that our method has excellent performance on diffusion-generated images, strong generalization ability, and excellent robustness.
\end{itemize}

\begin{figure*}
  \centering
  \includegraphics[width=1.0\textwidth]
  {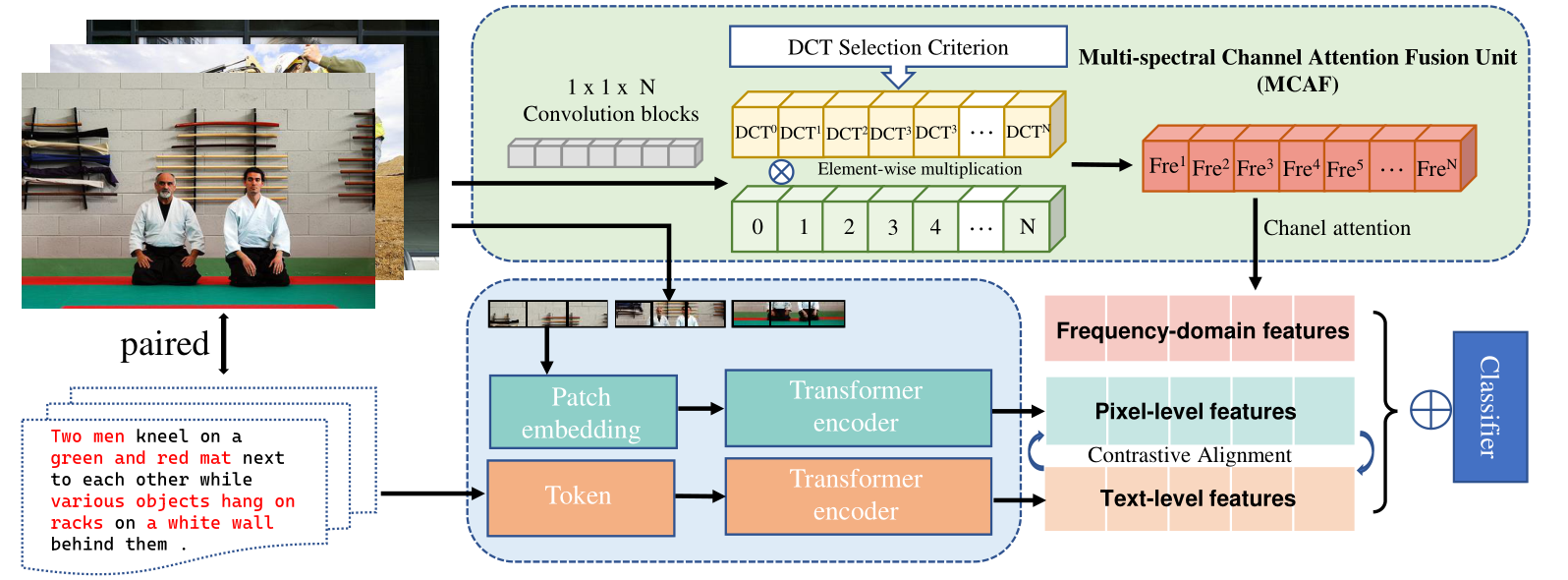} 
  \caption{\textbf{An illustration of diffusion—generated images detection.} The Text-Image Alignment and Extraction Module processes textual and visual information pairs, extracting aligned content information. Within the SpectraFuse Unit, DCT vectors extracted through the DCT Selection Criterion are applied to perform DCT transformation on individual channels of the image. Subsequently, a channel attention mechanism is employed to fuse frequency domain information.}
  \label{fig:model}
\end{figure*}

\section{Related work}

\subsection{Diffusion Model Image Generation}
Inspired by nonequilibrium thermodynamics, Ho et al.~\cite{ho_denoising_2020} proposed a new generation of paradigms, namely Denoised Diffusion Probabilistic Models (DDPMs), which achieved competitive performance compared with PGGAN~\cite{karras2018progressive} right out of the gate. Since then, more and more researchers have turned their attention to diffusion models. Song et al.~\cite{song_denoising_2022} generalized DDPM to Denoising Diffusion Implicit Models (DDIMs) through a class of non-Markovian diffusion processes, which resulted in more high-quality samples with fewer sampling steps. Later work ADM~\cite{dhariwal_diffusion_2021} found a more efficient architecture that further achieves state-of-the-art performance compared to other generative models with classifier bootstrapping.LDM~\cite{rombach_high-resolution_2022} modulates the input diffusion model through a cross-attention mechanism and proposes introducing latent diffusion models with latent space. The recently popular Stable Diffusion v1 and v2 are based on the LDM and have been further improved to achieve surprising performance in text-to-image generation. DALLE and DALLE2, released in the same period, utilize a priori models to generate image embedding of the input text based on CLIP~\cite{radford2021learning} and then generate images from this embedding by using a diffusion-based decoder.

\subsection{Fake Image Detection Techniques}

Generative image detection has been extensively studied in the last few years. Early researchers focused on the subtle changes in the tampering boundaries of the forged image to determine the authenticity of the image based on boundary artifacts and changes in statistical features~\cite{liu_deep_2018}~\cite{huh_fighting_2018}, wang et al.~\cite{wang_fakespotter_2020} used monitoring neurons to forensically examine GAN generated images. However, they neglected the ability to generalize for invisible generative models, and there are frequency-based methods in addition to spatial artifact detection. Subsequently, Frank et al.~\cite{frank_leveraging_2020} proposed that in the frequency domain, GAN has a special texture due to the presence of up-sampling operations in the generator when generating images. However, with the rapid development of diffusion modeling, a general and robust detector for detecting diffusion model-generated images has yet to be developed. We note that the problem of diffusion-generated image detection has also been noted in several recent works, e.g., as it was pointed out that the lack of explicit 3D modeling of objects and surfaces leads to asymmetry in shadows and reflection images~\cite{farid_lighting_2022}. There are also some works based on pre-trained models of text images for detection, but they do not propose a new paradigm for the development of forged image detection techniques, and they are simply a simple exploitation of several existing works~\cite{sha2023fake}. There are also papers that perform detection by measuring the error between the input image and the image reconstructed by a pre-trained Diffusion model~\cite{wang_dire_2023}, but this performance may be affected by the Diffusion model because different Diffusion models may produce different reconstruction error sizes, and it is necessary to propose some more generalized methods considering the current proliferation of Diffusion models. Unlike them, our work focuses on exploring a generalizable detector that can be applied to large-scale diffusion models.

\section{Method}
In this paper, we present a novel method named Trinity Detector for diffusion-generated image detection. The rest of this section is organized as follows. First, we briefly review DDPM and DCT. Then, we present details of Trinity Detector for diffusion-generated image detection. Finally, we introduce a new dataset, i.e., the diffusion-generated graphic pair dataset, for training and evaluating diffusion-generated image detectors.

\subsection{Preliminaries}
\subsubsection{Denoising Diffusion Probabilistic Models (DDPMs)}
DDPMs generate images by simulating a random diffusion process involving two Markov chains: a forward chain that perturbs data into noise and a reverse chain that transforms noise back into data. The forward chain is typically manually designed to convert any data distribution into a simple prior distribution, while the Markov chain of the reverse chain is learned by parameterizing the transition kernel with deep neural networks to invert the former. By initially sampling a random vector from a prior distribution and subsequently performing ancestral sampling through a reverse Markov chain, new samples are generated.

In the forward chain process, the data distribution \(x_0 \sim q(x_0)\) is transformed using the transition kernel \(q(x_t \mid x_{t-1})\) through a Markov process that defined as:
\begin{equation} \label{eq:markov_chain}
q(x_1, \ldots, x_T \mid x_0) = \prod_{t=1}^{T} q(x_t \mid x_{t-1}), \tag{1}
\end{equation}

In DDPMs, this transformation kernel is classically designed by adding a Gaussian perturbation by sampling the Gaussian vector $\epsilon \sim \mathcal{N}(0, I)$ and applying the transformation. Finally, $x_t$ is approximated as a Gaussian distribution, i.e.,  $x_t \sim q(x_T) \quad: \int q(x_T \mid x_0)q(x_0) \,dx_0 \approx \mathcal{N}(x_T; 0, I)$

The reverse process is also characterized as a Markov chain:
\begin{equation} \label{eq:markov_chain_theta}
p_{\theta}(x_{t-1} \mid x_t) = \mathcal{N}(x_{t-1}; \mu_{\theta}(x_t, t), \Sigma_{\theta}(x_t, t)), \tag{2}
\end{equation}

Diffusion models leverage a network, $\Phi(\cdot)$, to model the true distribution $p_{\theta}(x_{t-1}\mid x_t)$, where $\theta$ represents model parameters. The corresponding approximate distribution $q(x_{t-1}\mid x_t)$ is typically parameterized by a neural network. The overarching simplified optimization objective entails a sampling and denoising process, articulated as follows:
\begin{equation} \label{eq:simple_loss}
L_{\text{sim}}(\theta) = \mathbb{E}_{t, x_0, \epsilon} \left[ \lVert \epsilon - \epsilon_{\theta}(\sqrt{\alpha t}x_0 + \sqrt{1 - \alpha t}\epsilon, t) \rVert^2 \right],\tag{3}
\end{equation}

\subsubsection{Discrete Cosine Transform (DCT)}
Typically, the basis functions for two-dimensional (2D) Discrete Cosine Transform (DCT) is:
\begin{equation} \label{eq:cosine_formula}
B_{i,j}^{h,w} = \cos\left(\frac{\pi h}{H} \left(i + \frac{1}{2}\right)\right) \cos\left(\frac{\pi w}{W} \left(j + \frac{1}{2}\right)\right), \tag{4}
\end{equation}

Thus, the two-dimensional DCT can be expressed as:
\begin{equation} \label{eq:f_2d}
f^{2d}_{h,w} = \sum_{i=0}^{H-1} \sum_{j=0}^{W-1} x^{2d}_{i,j} B^{h,w}_{i,j}, \tag{5}
\end{equation}
$s.t. h \in \{0, 1, \ldots, H-1\}$, $w \in \{0, 1, \ldots, W-1\}$

Here, $\mathbf{f}^{2d} \in \mathbb{R}^{H \times W}$ represents the 2D DCT spectrum, $\mathbf{x}^{2d} \in \mathbb{R}^{H \times W}$ is the input, $H$ is the height of $\mathbf{x}^{2d}$, and $W$ is the width of $\mathbf{x}^{2d}$. Correspondingly, the inverse of the 2D DCT can be expressed as:
\begin{equation} \label{eq:inverse_dct}
x^{2d}_{i,j} = \frac{1}{H} \sum_{h=0}^{H-1} \sum_{w=0}^{W-1} f^{2d}_{h,w}  B^{h,w}_{i,j}, \tag{6}
\end{equation}
$s.t. i \in \{0, 1, \ldots, H-1\}, j \in \{0, 1, \ldots, W-1\}.$

\subsection{Trinity Detector}
We find that the images generated by the diffusion model are significantly different from the real images in terms of frequency domain information distribution compared to the real images. To address this, Trinity Detector leverages a Multi-spectral Channel Attention Fusion Unit (MCAF) that utilizes channel attention to model and process channels in the frequency domain. This enhances the representational capacity for frequency domain features, which are then fused with the image-text features extracted by a pre-trained encoder. Building upon these enhancements, Trinity Detector imparts discriminative properties for distinguishing diffusion-generated images from real images. Overall, our detection process is shown in Eq~\ref{eq:all}, where $\widetilde{\phi}_{\text{Text}}$ and $\widetilde{\phi}_{\text{Image}}$ denote text image extraction and alignment in semantic space, and $\phi_{\text{Frequency}}$ is adaptive fusion based on the channel attention mechanism to extract multi-band frequency domain information

\begin{equation}\label{eq:all}
\begin{split}
    \text{Feature} = & \widetilde{\phi}_{\text{Text}}(\text{Text}) \oplus \widetilde{\phi}_{\text{Image}}(\text{Image}) \\
    & \oplus \text{Attention}_{\text{Channel}}(\phi_{\text{Frequency}}(\text{Image})),
\end{split}
\tag{7}
\end{equation}

\textbf{MCAF} To enhance channel compression and introduce more information, we propose the Multi-spectral Channel Attention Fusion Unit (MCAF), which extends frequency extraction to more frequency components of the 2D DCT, compressing additional information from multiple frequency components of the 2D DCT.

Initially, the input image is divided into multiple parts along the channel dimension through convolution, denoted as $[X_0, X_1, \ldots, X_{n-1}]$, $X_i \in \mathbb{R}^{C' \times H \times W}, \quad i \in \{0, 1, \ldots, n-1\}$ For each part, a 2D DCT transformation is applied to the selected frequency components determined by the DCT Selection Criterion. This serves as the frequency domain channel attention, represented as follows:
\begin{equation}
\begin{split}
   Freq_i&= \sum_{h=0}^{H-1} \sum_{w=0}^{W-1} X^{i}_{:,h,w} B^{u_i,v_i}_{h,w} \\
    &\quad \text{s.t. } i \in \{0, 1, \ldots, n-1\}
\end{split}, 
\tag{8}
\end{equation}

The overall SpectraFuse process can be expressed as:

\begin{equation}
\begin{split}
Freq & = FreqAttention\{cat([Freq_0, \ldots, Freq_n])\} \\
& = sigmoid(fc(cat([Freq_0, \ldots, Freq_n])))
\end{split}, 
\tag{9}
\end{equation}

Where the entire process can be seen as the extraction and compression of multi-channel frequency domain features, our approach extends the original methods focused on specific frequency domains to a framework that encompasses multiple frequency components.

\textbf{DCT Selection Criterion} To implement multi-spectrum channel attention, we adopted the criteria proposed by Qin et al.~\cite{qin_fcanet_2021}, which include FcaNet-LF (Low-Frequency), FcaNet-TS (Two-Step Selection), and FcaNet-NAS (Neural Architecture Search). The optimal frequency components for channel attention are searched through neural network exploration. The frequency components for this part can be expressed as:
\begin{equation}
Freq^i_{nas} = \sum_{(u,v) \in O}\frac{ \exp(\alpha(u,v))}{\sum_{(u',v') \in O} \exp(\alpha(u',v'))} DCT^{u,v}_{2D}(X_i), \tag{10}
\end{equation}

\textbf{Image-Text Content Extraction Module} Combining the formidable text-image feature extraction and alignment capabilities demonstrated by CLIP, our approach strategically employs its pretrained text-image encoder. Leveraging this pretrained encoder facilitates the deep extraction and fusion of image features and semantic information. Specifically, we use Vit-32 as the image encoder and a transformer encoder for text, creating a cohesive multimodal representation.

This combined strategy not only enhances feature extraction but also promotes a deeper integration of semantic information, contributing to the improved effectiveness and versatility of our proposed method.

\subsection{TxtDiffusionForensics: Dataset for Evaluating Diffusion-Generated Image Detectors}

Due to the lack of public datasets for generating fake images from diffusion models, this paper produces the TxtDiffusionForensics dataset, which consists of a diffusion model and the corresponding text pairs. We select text cues from the Flickr30K and MSCOCO public datasets and use the stable diffusion model and the GLIDE model to generate synthetic images.

\section{Experiment}
In this section, we first describe the experimental setup and then provide extensive experimental results to demonstrate the superiority of our approach.
\subsection{Experimental Setup}
\hspace{0.5cm}\textbf{Data Pre-processing}
All experiments were conducted on our TxtDiffusionForensics dataset. We randomly selected 5000 real images and 5000 synthesized images, each paired with corresponding textual prompts, for the training dataset. The evaluation test set for assessing the performance of the hybrid detection model consists of three parts. One dataset is generated by StyleGAN, while the other two are generated by stable diffusion and GLIDE models, respectively. In assessing model performance, we employ classification accuracy (ACC).

\begin{table*}
\centering
\caption{Comparison with other detectors. We performed white-box testing on the StableDiffusion dataset and transferability black-box experiments on the other two datasets and investigate the robustness tests with different JPEG compression rates (80\%, 50\%), different Gaussian blurring rates ($\alpha$ = 1, 2), * denotes re-training on the Stable Diffusion dataset of TxtDiffusionForesnsic, Ori denotes the original image.}
\label{tab:com}
\resizebox{\linewidth}{!}{%
\begin{tabular}{>{\centering\hspace{0pt}}m{0.12\linewidth}|>{\centering\hspace{0pt}}m{0.05\linewidth}|>{\centering\hspace{0pt}}m{0.062\linewidth}>{\centering\hspace{0pt}}m{0.062\linewidth}|>{\centering\hspace{0pt}}m{0.054\linewidth}>{\centering\hspace{0pt}}m{0.054\linewidth}|>{\centering\hspace{0pt}}m{0.05\linewidth}|>{\centering\hspace{0pt}}m{0.062\linewidth}>{\centering\hspace{0pt}}m{0.062\linewidth}|>{\centering\hspace{0pt}}m{0.056\linewidth}>{\centering\hspace{0pt}}m{0.056\linewidth}|>{\centering\hspace{0pt}}m{0.05\linewidth}|>{\centering\hspace{0pt}}m{0.062\linewidth}>{\centering\hspace{0pt}}m{0.062\linewidth}|>{\centering\hspace{0pt}}m{0.056\linewidth}>{\centering\arraybackslash\hspace{0pt}}m{0.056\linewidth}} 
\hline
                                                      & \multicolumn{5}{>{\centering\hspace{0pt}}m{0.282\linewidth}|}{\underline{StableDiffusion}}                                                                                        & \multicolumn{5}{>{\centering\hspace{0pt}}m{0.286\linewidth}|}{GLIDE}                                                                                                    & \multicolumn{5}{>{\centering\arraybackslash\hspace{0pt}}m{0.286\linewidth}}{StyleGAN}                                                                                                  \\ 
\cline{2-16}
\multicolumn{1}{>{\hspace{0pt}}m{0.079\linewidth}|}{} & Ori            & \multicolumn{2}{>{\centering\hspace{0pt}}m{0.124\linewidth}|}{JEPG} & \multicolumn{2}{>{\centering\hspace{0pt}}m{0.108\linewidth}|}{Gaussian~} & Ori            & \multicolumn{2}{>{\centering\hspace{0pt}}m{0.124\linewidth}|}{JEPG} & \multicolumn{2}{>{\centering\hspace{0pt}}m{0.112\linewidth}|}{Gaussian~ ~} & Ori            & \multicolumn{2}{>{\centering\hspace{0pt}}m{0.124\linewidth}|}{JEPG} & \multicolumn{2}{>{\centering\arraybackslash\hspace{0pt}}m{0.112\linewidth}}{Gaussian~ ~}  \\ 
\hline
Method                                                &                & 80\%           & 50\%                                               & 1              & 2                                                             &                & 80\%           & 50\%                                               & 1              & 2                                                               &                & 80\%           & 50\%                                               & 1              & 2                                                                              \\ 
\hline
CNN\cite{wang_cnn-generated_2020}                                              & 0.493          & 0.534          & 0.387                                              & 0.456          & 0.389                                                         & 0.645          & 0.674          & 0.612                                              & 0.681          & 0.616                                                           & 0.827          & 0.707          & 0.611                                              & 0.714          & 0.598                                                                          \\
ProGAN\cite{gragnaniello_are_2021}                                           & 0.486          & 0.412          & 0.382                                              & 0.543          & 0.473                                                         & 0.503          & 0.436          & 0.427                                              & 0.459          & 0.429                                                           & 0.968          & 0.768          & 0.746                                              & 0.783          & 0.719                                                                          \\
StyleGAN\cite{gragnaniello_are_2021}                                        & 0.501          & 0.489          & 0.413                                              & 0.528          & 0.474                                                         & 0.510          & 0.542          & 0.436                                              & 0.536          & 0.438                                                           & \textbf{1.000} & \textbf{0.820} & 0.715                                              & \textbf{0.881} & 0.746                                                                          \\
CNN*\cite{wang_cnn-generated_2020}                                              & 0.931          & 0.781          & 0.616                                              & 0.733          & 0.625                                                         & 0.891          & 0.629          & 0.574                                              & 0.635          & 0.543                                                           & 0.764          & 0.636          & 0.539                                              & 0.693          & 0.510                                                                          \\
GAN*\cite{gragnaniello_are_2021}                                              & 0.943          & 0.757          & 0.599                                              & 0.749          & 0.578                                                         & 0.837          & 0.644          & 0.519                                              & 0.609          & 0.529                                                           & 0.913          & 0.811          & 0.730                                              & 0.837          & 0.694                                                                          \\
DE-FAKE\cite{sha2023fake}                                         & 0.853          & 0.748          & 0.682                                              & 0.767          & 0.739                                                         & 0.703          & 0.651          & 0.617                                              & 0.737          & 0.675                                                           & 0.657          & 0.607          & 0.544                                              & 0.611          & 0.537                                                                          \\
Dire\cite{wang_dire_2023}                                              & 0.977          & 0.809          & 0.718                                              & 0.902          & 0.856                                                         & 0.929          & 0.855          & 0.702                                              & 0.862          & 0.821                                                           & 0.835          & 0.720          & 0.689                                              & 0.785          & 0.697                                                                          \\ 
\hline
OURS                                                  & \textbf{0.993} & \textbf{0.931} & \textbf{0.894}                                     & \textbf{0.957} & \textbf{0.895}                                                & \textbf{0.945} & \textbf{0.868} & \textbf{0.790}                                     & \textbf{0.889} & \textbf{0.845}                                                  & 0.841          & 0.815          & \textbf{0.779}                                     & 0.806          & \textbf{0.755}                                                                 \\
\hline
\end{tabular}
}
\end{table*}

\textbf{Baselines}
To validate the exceptional performance of our proposed method, we have chosen several state-of-the-art methods that have achieved outstanding results on diverse datasets for comparison:

1.	CNNDetection~\cite{wang_cnn-generated_2020}: This method introduces an image detection model generated by a convolutional neural network (CNN). The model undergoes training on a specific CNN dataset and demonstrates the ability to generalize to other images generated by CNNs.

2.	GANDetection~\cite{gragnaniello_are_2021}: By training on ProGAN and StyleGAN, this method has achieved notable success in terms of generalization.

3.	DiffusionDetection: DIRE~\cite{wang_dire_2023} employs a pre-trained diffusion model to measure the error between input images and reconstructed images. DE-FAKE~\cite{sha2023fake} combines graphic and text content through the CLIP pre-trained model, and then classifies it on the classifier.

\subsection{Comparison to Existing Detectors}
In this study, we employed pre-trained weights obtained from official repositories to assess the performance of CNNDetection, GANDetection, and DiffusionDetection on our curated dataset. Since the DE-FAKE method is not yet open source, we follow the method in the original article to reproduce it on our dataset.

Quantitative results are detailed in Table~\ref{tab:com}. Notably, existing detectors exhibited a significant decline in performance when tasked with handling diffusion-generated images, with an accuracy (ACC) falling below 60\%. To address this limitation, we utilized diffusion-generated images as additional training data and conducted a retraining of CNNDetection and GANDetection. The resulting models demonstrated substantial improvements in detecting images generated by the same diffusion model used during training. However, their performance remained suboptimal when confronted with diffusion models not encountered during training.In contrast, our proposed method, Trinity Detector, showcases outstanding generalization capabilities.

\subsection{Robustness Assessment}
In addition to generalization to unknown generative models, robustness to unknown perturbations is also a general concern since, in practice, images are usually perturbed by various kinds of degradations. Here, we evaluate the robustness of the detector in two types of perturbations (i.e., Gaussian blur and JPEG compression). Perturbations are added in Gaussian blur ($\sigma = 1, 2$) and JPEG compression (quality = 80\%, 50\%). We explored the robustness of the Baseline and our Trinity Detector. The results are shown in Table~\ref{tab:com}. We observe that our DIRE achieves better performance at each level of Gaussian blurring and JPEG compression.

\subsection{Ablation Study} 
\begin{table}
\centering
\caption{The ablation experiment data for each module.}
\label{tab:abl}
\footnotesize
\begin{tblr}{
  cells = {c},
  hline{1-2,6} = {-}{},
}
          & Stable Diffusion             & GLIDE          & StyleGAN      \\
Trinity Detector      & \textbf{0.973} & \textbf{0.945} & \textbf{0.841} \\
Fre\textsubscript{ab}   & 0.676          & 0.559          & 0.513          \\
Caption\textsubscript{ab}   & 0.783          & 0.751          & 0.647          \\
Caption\textsubscript{BLIP} & 0.968          & 0.937          & 0.903          
\end{tblr}
\end{table}
Based on the description in Section 3.2, we introduce the Multi-spectral Channel Attention Fusion Unit (MCAF). In this section, we first perform a detailed ablation analysis of the module, evaluating and comparing its performance by training detectors that consider only text and image content. Considering that not all forged images contain textual descriptions, we performed ablation experiments with the text extraction unit and the corresponding experiments in the case of text generation using BLIP. The specific results are shown in Table~\ref{tab:abl}, and the Trinity Detector performs more superiorly compared to all the ablation detectors. Meanwhile, regardless of whether the text is a natural language text or a BLIP-generated text, it presents better results than the detector without text.

This indicates that combining frequency domain information with coarse-grained text and fine-grained visual content can effectively amplify the difference between the forged image and the real image, thus improving the detection of the forged image generated by the diffusion model.

\section{Conclusion}
In this paper, our goal is to develop a generalized detector to distinguish images generated by diffusion models. To address this challenge, we introduce the Trinity detector method. This approach is based on the observation that images generated by the diffusion model exhibit significant defects in the frequency domain. Therefore, we introduce a multispectral channel attention fusion unit, which adaptively fuses different frequency bands of real and diffusion model-generated images through an adaptive channel attention mechanism to extract their spectral inconsistencies in order to distinguish between real and faked images. Extensive experiments have verified that the proposed Trinity Detector method has better detection performance and robustness as well as generalization compared to other methods in detecting images generated from diffusion models..
% -------------------------------------------------------------------------
\bibliographystyle{IEEEbib}
\bibliography{icme2023template}
\end{document}